%% file: DVPF_FR_main.tex
\documentclass{article}
\usepackage{spconf}

\input{preamble}

\title{Deep Variational Privacy Funnel:\\General Modeling with Applications in Face Recognition
}

\name{Behrooz~Razeghi$^{\star}$ \qquad Parsa Rahimi$^{\star \dagger}$ \qquad S\'{e}bastien Marcel$^{ \star \ddagger}$%
	\thanks{This research is supported by the Swiss Center for Biometrics Research and Testing.}%
    \thanks{For the source code visit: \url{https://gitlab.idiap.ch/biometric/icassp2024.dvpf}.}%
}
\address{$^{\star}$ Idiap Research Institute \\
         $^{\dagger}$ École Polytechnique Fédérale de Lausanne (EPFL)\\
         $^{\ddagger}$ Université de Lausanne (UNIL)}

\begin{document}

\maketitle

\begin{abstract}
In this study, we harness the information-theoretic Privacy Funnel (PF) model to develop a method for privacy-preserving representation learning using an end-to-end training framework. We rigorously address the trade-off between obfuscation and utility. Both are quantified through the \textit{logarithmic loss}, a measure also recognized as self-information loss. This exploration deepens the interplay between information-theoretic privacy and representation learning, offering substantive insights into data protection mechanisms for both discriminative and generative models. Importantly, we apply our model to state-of-the-art face recognition systems. The model demonstrates adaptability across diverse inputs, from raw facial images to both derived or refined embeddings, and is competent in tasks such as classification, reconstruction, and generation. 

\vspace{-3pt}

\end{abstract}

\setlength{\abovedisplayskip}{7pt}
\setlength{\belowdisplayskip}{3pt}

\vspace{-7pt}

%
%
%
\section{Introduction}
\label{Sec:Introduction}

\vspace{-7pt}

In the fields of information theory and computer science, privacy preservation has been a perennial concern, evolving with technology and emerging privacy threats. The advent of big data intensified both the opportunities, such as innovative business models and personalized services, and challenges, including new privacy threats. Current privacy research pivots around a delicate balance between the provable privacy level and maintaining data utility, which may vary significantly depending on the specific application and data properties.

There exist two main types of privacy-preserving mechanisms: `\textit{prior-independent}' and `\textit{prior-dependent}'. Prior-independent mechanisms make minimal assumptions about the data distribution and adversary information, while prior-dependent mechanisms exploit knowledge about the data distribution and the adversary to protect privacy. Anonymization techniques like $k$-anonymity \cite{sweeney2002k}, $\ell$-diversity \cite{machanavajjhala2006diversity}, $t$-closeness \cite{li2007t}, differential privacy (DP) \cite{dwork2006calibrating}, and pufferfish \cite{kifer2012rigorous} aim to preserve data privacy by perturbing data. DP, in particular, is a widely used prior-independent metric that ensures statistical queries' results remain approximately the same regardless of the inclusion of an individual record in the dataset. 
Conversely, IT privacy \cite{calmon2015fundamental, kalantari2017information, osia2018deep, tripathy2019privacy, sreekumar2019optimal, diaz2019robustness, razeghi2020perfectobfuscation, rassouli2019optimal, atashin2021variational} works on designing mechanisms and metrics that preserve privacy when the statistical properties of the data are partially known or estimated. IT privacy approaches use metrics like $f$-divergences and Renyi divergence to model the trade-off between privacy (obfuscation) and utility, helping to understand the fundamental privacy limits.

Data-driven privacy mechanisms, like Generative Adversarial Networks (GANs) \cite{goodfellow2014generative} inspired ones, model the obfuscation-utility trade-off as a game between a defender (privatizer) and an adversary \cite{edwards2016censoring, hamm2017enhancing, huang2017context}. With the continuous improvement in machine learning capabilities, the importance of data-driven privacy mechanisms will increase. Privacy breaches can have serious consequences, hence the need to develop robust privacy-preserving techniques to protect sensitive information.

%
%

\vspace{5pt}

The primary contributions of our work are as follows:\vspace{-5pt}
\begin{itemize}[leftmargin=1em]
\item
%
To the best of our knowledge, ours is among the first comprehensive studies on Privacy Funnel (PF) modeling within the domain of deep learning. 
We establish a connection between the information-theoretic foundations of privacy and privacy-preserving representation learning, with a particular emphasis on cutting-edge face recognition systems.
\vspace{-7pt}
\item
We introduce a tight variational bound for information leakage which sheds light on the complexities inherent in privacy preservation during deep variational PF (DVPF) learning.\vspace{-7pt}
\item
Our insights into the upper bound of information leakage play as a crucial role in guiding the optimization of privacy-preserving synthetic data generation techniques.\vspace{-7pt}
\item
Our model is proficient in processing both raw image samples and facial image-derived embeddings. Its versatility spans classification, reconstruction, and generation tasks, and its inherent robustness distinguishes it. In alignment with our commitment to furthering research, a comprehensive package will be released, with its particulars detailed in the extended version of our paper.
\end{itemize}

%
\section{Privacy Funnel Model}

\vspace{-5pt}

Consider two correlated random variables $\mathbf{S}$ and $\mathbf{X}$ with a joint distribution $P_{\mathbf{S,X}}$. The objective of the Privacy Funnel (PF) method \cite{makhdoumi2014information} is to derive a representation $\mathbf{Z}$ of $\mathbf{X}$ through a stochastic mapping $P_{\mathbf{Z}\mid \mathbf{X}}$, satisfying the following conditions:
(i) $\mathbf{S} \markov \mathbf{X} \markov \mathbf{Z}$,
(ii) representation $\mathbf{Z}$ maximizes the mutual information about $\mathbf{X}$ (i.e., $\I \left( \mathbf{X}; \mathbf{Z} \right)$), and
(iii) representation $\mathbf{Z}$ minimizes the mutual information about $\mathbf{S}$ (i.e., $\I \left( \mathbf{S}; \mathbf{Z}\right)$). 
In essence, the PF method meticulously navigates the balance between the potential information leakage, $\I \left( \mathbf{S}; \mathbf{Z} \right)$, and the utility of the revealed information, $\I \left( \mathbf{X}; \mathbf{Z} \right)$.
The functional representation of the Privacy Funnel can be expressed as:\vspace{-5pt}
\begin{equation}\label{Eq:PF_LagrangianFunctional}
\!\!\!\! \mathsf{PF} \left( R^{\mathrm{s}}, P_{\mathbf{S}, \mathbf{X}}\right)  \coloneqq \!\!\!\!\!\! \mathop{\sup}_{\substack{P_{\mathbf{Z} \mid \mathbf{X}}: \\  \mathbf{S} \markov \mathbf{X} \markov \mathbf{Z}}} \!\!\!\! \I \left( \mathbf{X}; \mathbf{Z} \right) \;\;  \mathrm{s.t.} \;\;  \I \left( \mathbf{S}; \mathbf{Z} \right) \leq R^{\mathrm{s}}. 
\end{equation}
The PF curve is defined by the values $\mathsf{PF} \left( R^{\mathrm{s}}, P_{\mathbf{S}, \mathbf{X}}\right)$ for different $R^{\mathrm{s}}$. We can use a Lagrange multiplier $\alpha \geq 0$ to represent the PF problem by the associated Lagrangian functional:
$\mathcal{L}_{\mathrm{PF}} \!\left( P_{\mathbf{Z}\mid \mathbf{X}}, \alpha \right)  \! \coloneqq  \! \I  \left( \mathbf{X}; \mathbf{Z} \right) - \alpha \,   \I \left( \mathbf{S}; \mathbf{Z} \right)$. 
Note that the PF model emerges as a specific instance of the CLUB model \cite{razeghi2023bottlenecks} when the utility information corresponds directly to data $\mathbf{X}$ and the information complexity of the CLUB model exceeds Shannon entropy $\H \left( P_{\mathbf{X}}\right)$.

%
%

\vspace{2pt}

Our threat model includes the following assumptions:\vspace{-5pt}
\begin{itemize}[leftmargin=1em]
    \item
    We consider an adversary who is interested in a specific attribute $\mathbf{S}$ related to the data $\mathbf{X}$. This attribute $\mathbf{S}$ could be any function of $\mathbf{X}$, possibly randomized. We restrict $\mathbf{S}$ to represent a discrete attribute, covering prevalent scenarios of interest, such as facial features or identity attributes.\vspace{-7pt}
    \item
    The adversary has access to the released representation $\mathbf{Z}$ and respects the Markov chain relationship $\mathbf{S} \markov \mathbf{X} \markov \mathbf{Z}$.\vspace{-7pt}
    \item
    The mapping $P_{\mathbf{Z} \mid \mathbf{X}}$, designed by the defender (privatizer), is assumed to be public knowledge. This implies that the adversary is aware of the strategy employed by the defender.
\end{itemize}

\vspace{-16pt}

%
\section{Deep Variational Privacy Funnel}
\label{GVPF}

\vspace{-8pt}

In this section, we introduce our core methodology, the Deep Variational Privacy Funnel (DVPF). Building on the PF principle, this framework utilizes deep neural networks to optimize the information obfuscation-utility trade-offs.

\vspace{-9pt}

\noindent
\subsection{Parameterized Variational Approximation of $\I (\mathbf{S}; \mathbf{Z})$}

\vspace{-5pt}
 
We provide parameterized variational approximations for information leakage, which include both an explicit tight variational bound and an upper bound. To better understand the nature of information leakage, we can express $\I \left( \mathbf{S}; \mathbf{Z} \right)$ as 
$\I \left( \mathbf{X}; \mathbf{Z} \right) - \I \left( \mathbf{X}; \mathbf{Z} \mid \mathbf{S} \right) 
 = 
\I \left( \mathbf{X}; \mathbf{Z} \right) - \H \left( \mathbf{X} \mid \mathbf{S} \right) + \H \left(  \mathbf{X} \mid \mathbf{S}, \mathbf{Z}\right).$
The conditional entropy $\H \left( \mathbf{X} \mid \mathbf{S} \right)$ is originated from the nature of data, since it is out of our control. Now, we derive the variational decomposition of $\I \left( \mathbf{X}; \mathbf{Z} \right)$ and $\H \left(  \mathbf{X} \mid \mathbf{S}, \mathbf{Z}\right)$. 
The mutual information $\I \left( \mathbf{X}; \mathbf{Z} \right)$ can be decomposed as:\vspace{-5pt}
\begin{equation}
\label{I_xz_decomposition}
\I \left( \mathbf{X}; \mathbf{Z} \right) =
\D_{\mathrm{KL}} \left( P_{\mathbf{Z} \mid \mathbf{X}} \Vert Q_{\mathbf{Z}} \mid P_{\mathbf{X}} \right) - \D_{\mathrm{KL}} \left( P_{\mathbf{Z}} \Vert Q_{\mathbf{Z}}\right),
\end{equation}
where $Q_{\mathbf{Z}} \!: \! \mathcal{Z} \! \rightarrow \! \mathcal{P}\left( \mathcal{Z}\right)$ is variational approximation of the latent space distribution $P_{\mathbf{Z}}$. 
The conditional entropy $\H \left( \mathbf{X} \! \mid \! \mathbf{S}, \mathbf{Z} \right)$ can be decomposed as:\vspace{-5pt}
\begin{subequations}
\label{ConditionalEntropy_X_given_SZ}
\begin{align}
\label{ConditionalEntropy_X_given_SZ_a}
& \H \left( \mathbf{X} \! \mid \! \mathbf{S}, \mathbf{Z} \right) \\
& = -\mathbb{E}_{P_{\mathbf{S}, \mathbf{X}}} \left[  \mathbb{E}_{P_{\mathbf{Z} \mid \mathbf{X}}} \left[ \log Q_{\mathbf{X} \mid \mathbf{S}, \mathbf{Z}} \right] \right] -  \D_{\mathrm{KL}} \left( P_{\mathbf{X}\mid \mathbf{S}, \mathbf{Z} }  \Vert Q_{\mathbf{X} \mid \mathbf{S}, \mathbf{Z}} \right)  \nonumber\\
& \leq -\mathbb{E}_{P_{\mathbf{S}, \mathbf{X}}} \left[  \mathbb{E}_{P_{\mathbf{Z} \mid \mathbf{X}}} \left[ \log Q_{\mathbf{X} \mid \mathbf{S}, \mathbf{Z}} \right] \right] \eqqcolon \H^{\mathrm{U}} \! \left( \mathbf{X} \! \mid \! \mathbf{S}, \mathbf{Z} \right), \label{ConditionalEntropy_X_given_SZ_b}
\end{align}
\end{subequations}
where $Q_{\mathbf{X}\mid \mathbf{S}, \mathbf{Z}}\! :\! \mathcal{S} \! \times \! \mathcal{Z} \! \rightarrow \! \mathcal{P}\left( \mathcal{X}\right)$ is variational approximation of the optimal uncertainty decoder distribution $P_{\mathbf{X}\mid \mathbf{S}, \mathbf{Z}}$, and the inequality in \eqref{ConditionalEntropy_X_given_SZ_b} follows by noticing that $ \D_{\mathrm{KL}}  ( P_{\mathbf{X}\mid \mathbf{S}, \mathbf{Z} }  \Vert Q_{\mathbf{X} \mid \mathbf{S}, \mathbf{Z}}  )$ $\geq 0$. 
Using \eqref{I_xz_decomposition} and \eqref{ConditionalEntropy_X_given_SZ}, the \textit{variational upper bound} of information leakage is given as:\vspace{-8pt}
\begin{multline}\label{I_SZ_upperBound}
    \I \left( \mathbf{S}; \mathbf{Z} \right) \leq \D_{\mathrm{KL}} \left( P_{\mathbf{Z} \mid \mathbf{X}} \Vert Q_{\mathbf{Z}} \mid P_{\mathbf{X}} \right) - \D_{\mathrm{KL}} \left( P_{\mathbf{Z}} \Vert Q_{\mathbf{Z}}\right)  \\ + \H^{\mathrm{U}} \! \left( \mathbf{X} \! \mid \! \mathbf{S}, \mathbf{Z} \right). 
\end{multline}

\vspace{-6pt}

We now employ neural networks to approximate the parameterized variational upper bound of information leakage. 
Let $P_{\boldsymbol{\phi}} (\mathbf{Z} \! \mid \! \mathbf{X})$ represent the family of encoding probability distributions $P_{\mathbf{Z} \mid \mathbf{X}}$ over $\mathcal{Z}$ for each element of space $\mathcal{X}$, parameterized by the output of a deep neural network $f_{\boldsymbol{\phi}}$ with parameters $\boldsymbol{\phi}$. 
Analogously, let $P_{\boldsymbol{\varphi}}  \left( \mathbf{X} \! \mid \! \mathbf{S}, \mathbf{Z} \right)$ denote the corresponding family of decoding probability distributions $Q_{\mathbf{X} \mid \mathbf{S}, \mathbf{Z}}$, driven by $g_{\boldsymbol{\varphi}}$. 
Lastly, $Q_{\boldsymbol{\psi}} (\mathbf{Z})$ denotes the parameterized prior distribution, either explicit or implicit, that is associated with $Q_{\mathbf{Z}}$. 
Using \eqref{I_xz_decomposition}, the parameterized variational approximation of $\I \left( \mathbf{X}; \mathbf{Z} \right)$ can be defined as:\vspace{-6pt}
\begin{multline}\label{Eq:I_XZ_phi_psi}
\I_{\boldsymbol{\phi}, \boldsymbol{\psi}}   \left( \mathbf{X}; \mathbf{Z} \right) \coloneqq  \D_{\mathrm{KL}} \! \left( P_{\boldsymbol{\phi}} (\mathbf{Z} \! \mid \! \mathbf{X}) \, \Vert \, Q_{\boldsymbol{\psi}}   (\mathbf{Z}) \mid P_{\mathsf{D}}(\mathbf{X}) \right)\\ -  \D_{\mathrm{KL}} \! \left( P_{\boldsymbol{\phi}}(\mathbf{Z}) \, \Vert \, Q_{\boldsymbol{\psi}} (\mathbf{Z})\right).
\end{multline}
The parameterized variational approximation of conditional entropy $\H^{\mathrm{U}} \left( \mathbf{X} \mid \mathbf{S}, \mathbf{Z} \right)$ in \eqref{ConditionalEntropy_X_given_SZ_b} can be defined as:\vspace{-5pt}
\begin{eqnarray}
 \H_{\boldsymbol{\phi}, \boldsymbol{\varphi}}^{\mathrm{U}} \! \left( \mathbf{X} \! \mid \! \mathbf{S}, \mathbf{Z} \right) \! \coloneqq \! - \mathbb{E}_{P_{\mathbf{S}, \mathbf{X}}} \! \left[  \mathbb{E}_{P_{\boldsymbol{\phi}} (\mathbf{Z} \mid \mathbf{X})} \! \left[ \log P_{\boldsymbol{\varphi}}  (\mathbf{X} \! \mid \! \mathbf{S}, \mathbf{Z}) \right] \right] .\!\!\!\!
\end{eqnarray}
Let $\I_{\boldsymbol{\phi}, \boldsymbol{\xi}} \left( \mathbf{S}; \mathbf{Z} \right)$ denote the parameterized variational approximation of information leakage $\I \left(\mathbf{S}; \mathbf{Z} \right)$.
Using \eqref{I_SZ_upperBound}, an upper bound of $\I_{\boldsymbol{\phi}, \boldsymbol{\xi}}  \! \left( \mathbf{S}; \mathbf{Z} \right)$ can be given as:\vspace{-5pt}
\begin{subequations}\label{Eq:I_SZ_phi_Xi_UpperBound}
\begin{align}
\!\!\!\!\! \I_{\boldsymbol{\phi}, \boldsymbol{\xi}} (\mathbf{S}; \mathbf{Z})  &\leq  \!\!\!\!\!\!\!\!\!\!
\underbrace{\I_{\boldsymbol{\phi}, \boldsymbol{\psi}}  \left( \mathbf{X}; \mathbf{Z} \right)}_{\mathrm{Information~Complexity}}  \!\! + \!  \underbrace{\H_{\boldsymbol{\phi}, \boldsymbol{\varphi}}^{\mathrm{U}}   \left( \mathbf{X} \! \mid \! \mathbf{S}, \mathbf{Z} \right)}_{\mathrm{Information~Uncertainty}} \!\!\!\!  + \, \mathrm{c} \\ 
&\eqqcolon \; \I_{\boldsymbol{\phi}, \boldsymbol{\psi}, \boldsymbol{\varphi}}^{\mathrm{U}}   \left( \mathbf{S}; \mathbf{Z} \right) + \mathrm{c},
\end{align}
\end{subequations}
where $\mathrm{c}$ is a constant term, independent of the neural networks parameters. 
This upper bound encourages the model to reduce both the information complexity, represented by $\I_{\boldsymbol{\phi}, \boldsymbol{\psi}}  \left( \mathbf{X}; \mathbf{Z} \right)$, and the information uncertainty, denoted by $\H_{\boldsymbol{\phi}, \boldsymbol{\varphi}}^{\mathrm{U}}  \left( \mathbf{X} \! \mid \! \mathbf{S}, \mathbf{Z} \right)$. Consequently, this leads the model to forget or de-emphasize the sensitive attribute $\mathbf{S}$, which subsequently reduces the uncertainty about the useful~data~$\mathbf{X}$. In essence, this nudges the model towards an accurate~reconstruction~of~the~data~$\mathbf{X}$.

Now, let us derive another parameterized variational bound of information leakage $\I_{\boldsymbol{\phi}, \boldsymbol{\xi}} \left( \mathbf{S}; \mathbf{Z} \right)$ \cite{razeghi2023bottlenecks}. 
We can decompose $\I_{\boldsymbol{\phi}, \boldsymbol{\xi}}  \left( \mathbf{S}; \mathbf{Z} \right)$ as follows:\vspace{-8pt}
\begin{align}\label{Eq:I_SZ_phi_xi_SecondDecomposition}
&\I_{\boldsymbol{\phi}, \boldsymbol{\xi}}  \left( \mathbf{S}; \mathbf{Z} \right)  \\
& =    
 \underbrace{- \H_{\boldsymbol{\phi}, \boldsymbol{\xi}} \left( \mathbf{S} \! \mid  \! \mathbf{Z} \right) 
+ \H \left( P_{\mathbf{S}} \, \Vert \, P_{\boldsymbol{\xi}} (\mathbf{S}) \right)}_{\mathrm{Prediction~Fidelity}}\,
- \!\!\!\!  \underbrace{ \D_{\mathrm{KL}} \left( P_{\mathbf{S}} \, \Vert \, P_{\boldsymbol{\xi}} (\mathbf{S}) \right)}_{\mathrm{Distribution~Discrepancy}},\nonumber
\end{align}
where $P_{\boldsymbol{\xi}} (\mathbf{S} \! \mid \! \mathbf{Z})$ denotes the corresponding family of decoding probability distribution $Q_{\mathbf{S} \mid \mathbf{Z}}$, where $Q_{\mathbf{S} \mid \mathbf{Z}} : \mathcal{Z} \rightarrow \mathcal{P} (\mathcal{S})$ is a variational approximation of optimal decoder distribution $P_{\mathbf{S} \mid \mathbf{Z}}$. 
This information decomposition encourages the model to (i) increase uncertainty regarding the sensitive attribute $\mathbf{S}$ upon knowing the released representation $\mathbf{Z}$. Specifically, the goal is to attain maximum entropy for a discrete sensitive attribute $\mathbf{S}$ when all conditional distributions are uniform. This means the adversary, lacking any additional information, can do no better than `\textit{random guessing}'. This scenario equates to a potential lower boundary for $- \H_{\boldsymbol{\phi}, \boldsymbol{\xi}} \left( \mathbf{S} \! \mid  \! \mathbf{Z} \right) $ at $- \log_2 N$ and upper boundary for $\H \left( P_{\mathbf{S}} \, \Vert \, P_{\boldsymbol{\xi}} (\mathbf{S}) \right)$ at $\log_2 N$, where $N$ represents the possible states (or values, or classes) of $\mathbf{S}$. 
(ii) Ensure the model's inferred distribution, $ P_{\boldsymbol{\xi}} (\mathbf{S})$, aligns tightly with the actual distribution $P_{\mathbf{S}}$. Ideally, the divergence measure, $\D_{\mathrm{KL}} \left( P_{\mathbf{S}}  \Vert  P_{\boldsymbol{\xi}} (\mathbf{S}) \right)$, is minimized to zero when $ P_{\boldsymbol{\xi}} (\mathbf{S})$ aligns perfectly with $P_{\mathbf{S}}$.
It's essential to recognize that, although the parameterized approximation in \eqref{Eq:I_SZ_phi_xi_SecondDecomposition} doesn't explicitly rely on the information complexity $\I_{\boldsymbol{\phi}, \boldsymbol{\psi}} \left( \mathbf{X}; \mathbf{Z} \right)$, it is intrinsically linked through the encoder $f_{\boldsymbol{\phi}}$.

\vspace{-12pt}
 
\subsection{Parameterized Variational Approximation of $\I(\mathbf{X}; \mathbf{Z})$} 

\vspace{-8pt}

We now quantify information utility by decomposing the mutual information $\I (\mathbf{X}; \mathbf{Z})$ and deriving its parameterized variational approximation. The end-to-end parameterized variational approximation associated to the information utility $\I (\mathbf{X}; \mathbf{Z})$ can be defined as:\vspace{-6pt}
\begin{subequations}\label{Eq:I_XZ_phi_theta}
\begin{align}
\!\!\! \I_{\boldsymbol{\phi}, \boldsymbol{\theta}}  \left( \mathbf{X}; \mathbf{Z} \right) 
& \!  \coloneqq  \!
\mathbb{E}_{P_{\mathsf{D}} (\mathbf{X})} \left[  \mathbb{E}_{P_{\boldsymbol{\phi}} \left( \mathbf{Z} \mid \mathbf{X} \right) } \left[ \log P_{\boldsymbol{\theta}}   \left( \mathbf{X} \! \mid \! \mathbf{Z} \right) \right] \right] \\
& - \D_{\mathrm{KL}} \left( P_{\mathsf{D}} (\mathbf{X}) \Vert P_{\boldsymbol{\theta}} (\mathbf{X}) \right)
+ \H \left( P_{\mathsf{D}} (\mathbf{X})  \Vert P_{\boldsymbol{\theta}} (\mathbf{X}) \right) 
  \nonumber \\
& \geq \!\!\!\!\!\!\!\!
  \underbrace{- \H_{\boldsymbol{\phi}, \boldsymbol{\theta}} \! \left( \mathbf{X} \! \mid  \! \mathbf{Z} \right)}_{\mathrm{Reconstruction~Fidelity}}
 \!\!\!\!\!\! -     \underbrace{\D_{\mathrm{KL}} \! \left( P_{\mathsf{D}} (\mathbf{X}) \Vert P_{\boldsymbol{\theta}} (\mathbf{X}) \right)}_{\mathrm{Distribution~Discrepancy}} \!\\
 & \eqqcolon \,\,\, \I_{\boldsymbol{\phi}, \boldsymbol{\theta}}^{\mathrm{L}}  \left( \mathbf{X}; \mathbf{Z} \right),
\end{align}
\end{subequations}
where $\H_{\boldsymbol{\phi}, \boldsymbol{\theta}} \left( \mathbf{X} \! \mid  \! \mathbf{Z} \right) \coloneqq \mathbb{E}_{P_{\mathsf{D}} (\mathbf{X})} \left[  \mathbb{E}_{P_{\boldsymbol{\phi}} \left( \mathbf{Z} \mid \mathbf{X} \right) } \left[ \log P_{\boldsymbol{\theta}} \left( \mathbf{X} \! \mid \! \mathbf{Z} \right) \right] \right]$.

\begin{table*}[h]
    \caption{Evaluation of facial recognition models using various backbones and loss functions. Metrics include entropy, mutual information between embeddings and labels (gender and race), and recognition accuracy on the `Morph' and `FairFace' datasets.}
    \vspace{-9pt}
    \label{Table:PerformanceMetrics_FacialRecognitionModels}
    \centering
    \resizebox{0.82\linewidth}{!}{%
    \begin{tabular}{cccc|cccccccccccc}
    \cline{5-16} 
\multicolumn{4}{c||}{}                                                                                           & \multicolumn{6}{c|}{$\mathbf{S}$: Gender}  & \multicolumn{6}{c|}{$\mathbf{S}$: Race}  \\ \cline{5-16}
\multicolumn{4}{c||}{}   & \multicolumn{2}{c|}{$\H(\mathbf{S})$} &\multicolumn{2}{c|}{$\I(\mathbf{X}; \mathbf{S})$} & \multicolumn{2}{c|}{Acc} & \multicolumn{2}{c|}{$\H(\mathbf{S})$} &\multicolumn{2}{c|}{$\I(\mathbf{X}; \mathbf{S})$} & \multicolumn{2}{c|}{Acc} \\ \hline
\rowcolor{headercolor}
\multicolumn{1}{|c|}{\begin{tabular}[c]{@{}c@{}}Backbone Dataset\end{tabular}} & \multicolumn{1}{c|}{Backbone}  & \multicolumn{1}{c|}{Loss Function} & \multicolumn{1}{c||}{Applied Dataset} & \multicolumn{1}{c|}{Train} & \multicolumn{1}{c|}{Test} & \multicolumn{1}{c|}{Train} & \multicolumn{1}{c|}{Test} & \multicolumn{1}{c|}{Train} & \multicolumn{1}{c|}{Test} & \multicolumn{1}{c|}{Train} & \multicolumn{1}{c|}{Test} & \multicolumn{1}{c|}{Train} & \multicolumn{1}{c|}{Test}  & \multicolumn{1}{c|}{Train} & \multicolumn{1}{c|}{Test}  \\ \hline \cline{1-16} 
\multicolumn{1}{|c|}{WebFace\textbf{4M}} & \multicolumn{1}{c|}{iresnet\textbf{18}} & \multicolumn{1}{c|}{AdaFace} & \multicolumn{1}{c||}{Morph} & \multicolumn{1}{c|}{\multirow{5}{*}{0.619}} &  \multicolumn{1}{c|}{\multirow{5}{*}{0.621}} & \multicolumn{1}{c|}{0.610}   & \multicolumn{1}{c|}{0.620}  & \multicolumn{1}{c|}{0.999}   & \multicolumn{1}{c|}{0.996}  & \multicolumn{1}{c|}{\multirow{5}{*}{0.924}} & \multicolumn{1}{c|}{\multirow{5}{*}{0.933}} & \multicolumn{1}{c|}{0.878}   & \multicolumn{1}{c|}{0.924}    & \multicolumn{1}{c|}{0.998}   & \multicolumn{1}{c|}{0.993} \\ \cline{1-4} \cline{7-10}  \cline{13-16} 
\rowcolor{lightgray}
\multicolumn{1}{|c|}{WebFace\textbf{4M}} & \multicolumn{1}{c|}{iresnet\textbf{50}} & \multicolumn{1}{c|}{AdaFace} & \multicolumn{1}{c||}{Morph} & \multicolumn{1}{c|}{\cellcolor{white}} & \multicolumn{1}{c|}{\cellcolor{white}} & \multicolumn{1}{c|}{0.610}   & \multicolumn{1}{c|}{0.620}    & \multicolumn{1}{c|}{0.999}   & \multicolumn{1}{c|}{0.996} & \multicolumn{1}{c|}{\cellcolor{white}} & \multicolumn{1}{c|}{\cellcolor{white}} & \multicolumn{1}{c|}{0.873}   & \multicolumn{1}{c|}{0.930}    & \multicolumn{1}{c|}{0.998}   & \multicolumn{1}{c|}{0.992}  \\ \cline{1-4} \cline{7-10}  \cline{13-16} 
\multicolumn{1}{|c|}{WebFace\textbf{12M}} & \multicolumn{1}{c|}{iresnet\textbf{101}} & \multicolumn{1}{c|}{AdaFace}  & \multicolumn{1}{c||}{Morph} & \multicolumn{1}{c|}{} & \multicolumn{1}{c|}{} & \multicolumn{1}{c|}{0.605}   & \multicolumn{1}{c|}{0.622}    & \multicolumn{1}{c|}{0.999}   & \multicolumn{1}{c|}{0.996} & \multicolumn{1}{c|}{} & \multicolumn{1}{c|}{} & \multicolumn{1}{c|}{0.873}   & \multicolumn{1}{c|}{0.911}    & \multicolumn{1}{c|}{0.998}   & \multicolumn{1}{c|}{0.992} \\ \cline{1-4} \cline{7-10}  \cline{13-16}
\rowcolor{lightgray}
\multicolumn{1}{|c|}{MS1M-RetinaFace} & \multicolumn{1}{c|}{iresnet\textbf{50}} & \multicolumn{1}{c|}{ArcFace} & \multicolumn{1}{c||}{Morph} & \multicolumn{1}{c|}{\cellcolor{white}} & \multicolumn{1}{c|}{\cellcolor{white}} & \multicolumn{1}{c|}{0.600}   & \multicolumn{1}{c|}{0.620}    & \multicolumn{1}{c|}{0.999}   & \multicolumn{1}{c|}{0.996} & \multicolumn{1}{c|}{\cellcolor{white}} & \multicolumn{1}{c|}{\cellcolor{white}} & \multicolumn{1}{c|}{0.865}   & \multicolumn{1}{c|}{0.910}    & \multicolumn{1}{c|}{0.997}   & \multicolumn{1}{c|}{0.993}  \\ \cline{1-4} \cline{7-10}  \cline{13-16}
\multicolumn{1}{|c|}{MS1M-RetinaFace} & \multicolumn{1}{c|}{iresnet\textbf{100}} & \multicolumn{1}{c|}{ArcFace} & \multicolumn{1}{c||}{Morph} & \multicolumn{1}{c|}{\cellcolor{white}} & \multicolumn{1}{c|}{\cellcolor{white}} & \multicolumn{1}{c|}{0.597}   & \multicolumn{1}{c|}{0.618}    & \multicolumn{1}{c|}{0.999}   & \multicolumn{1}{c|}{0.997} & \multicolumn{1}{c|}{\cellcolor{white}} & \multicolumn{1}{c|}{\cellcolor{white}} & \multicolumn{1}{c|}{0.868}   & \multicolumn{1}{c|}{0.905}    & \multicolumn{1}{c|}{0.997}   & \multicolumn{1}{c|}{0.993}  \\ \cline{1-16} 
\rowcolor{lightgray}
\multicolumn{1}{|c|}{WebFace\textbf{4M}} & \multicolumn{1}{c|}{iresnet\textbf{18}} & \multicolumn{1}{c|}{AdaFace} & \multicolumn{1}{c||}{FairFace} & \multicolumn{1}{c|}{\cellcolor{white}\multirow{5}{*}{0.999}} &  \multicolumn{1}{c|}{\cellcolor{white}\multirow{5}{*}{0.999}} & \multicolumn{1}{c|}{0.930}   & \multicolumn{1}{c|}{0.968}  & \multicolumn{1}{c|}{0.953}   & \multicolumn{1}{c|}{0.923}  & \multicolumn{1}{c|}{\cellcolor{white}\multirow{5}{*}{2.517}} & \multicolumn{1}{c|}{\cellcolor{white}\multirow{5}{*}{2.515}} & \multicolumn{1}{c|}{2.099}   & \multicolumn{1}{c|}{2.405}    & \multicolumn{1}{c|}{0.882}   & \multicolumn{1}{c|}{0.763} \\ \cline{1-4} \cline{7-10}  \cline{13-16} 
\multicolumn{1}{|c|}{WebFace\textbf{4M}} & \multicolumn{1}{c|}{iresnet\textbf{50}} & \multicolumn{1}{c|}{AdaFace} & \multicolumn{1}{c||}{FairFace} & \multicolumn{1}{c|}{} & \multicolumn{1}{c|}{} & \multicolumn{1}{c|}{0.932}   & \multicolumn{1}{c|}{0.968}    & \multicolumn{1}{c|}{0.954}   & \multicolumn{1}{c|}{0.931} & \multicolumn{1}{c|}{} & \multicolumn{1}{c|}{} & \multicolumn{1}{c|}{2.113}   & \multicolumn{1}{c|}{2.409}    & \multicolumn{1}{c|}{0.883}   & \multicolumn{1}{c|}{0.769}  \\ \cline{1-4} \cline{7-10}  \cline{13-16} 
\multicolumn{1}{|c|}{\cellcolor{lightgray}WebFace\textbf{12M}} & \multicolumn{1}{c|}{\cellcolor{lightgray}iresnet\textbf{101}} & \multicolumn{1}{c|}{\cellcolor{lightgray}AdaFace}  & \multicolumn{1}{c||}{\cellcolor{lightgray}FairFace} & \multicolumn{1}{c|}{} & \multicolumn{1}{c|}{} & \multicolumn{1}{c|}{\cellcolor{lightgray}0.934}   & \multicolumn{1}{c|}{\cellcolor{lightgray}0.969}    & \multicolumn{1}{c|}{\cellcolor{lightgray}0.957}   & \multicolumn{1}{c|}{\cellcolor{lightgray}0.930} & \multicolumn{1}{c|}{} & \multicolumn{1}{c|}{} & \multicolumn{1}{c|}{\cellcolor{lightgray}2.151}   & \multicolumn{1}{c|}{\cellcolor{lightgray}2.417}    & \multicolumn{1}{c|}{\cellcolor{lightgray}0.892}   & \multicolumn{1}{c|}{\cellcolor{lightgray}0.765}    \\ \cline{1-4} \cline{7-10}  \cline{13-16} 
\multicolumn{1}{|c|}{MS1M-RetinaFace} & \multicolumn{1}{c|}{iresnet\textbf{50}} & \multicolumn{1}{c|}{ArcFace} & \multicolumn{1}{c||}{FairFace} & \multicolumn{1}{c|}{\cellcolor{white}} & \multicolumn{1}{c|}{\cellcolor{white}} & \multicolumn{1}{c|}{0.892}   & \multicolumn{1}{c|}{0.962}    & \multicolumn{1}{c|}{0.950}   & \multicolumn{1}{c|}{0.927} & \multicolumn{1}{c|}{\cellcolor{white}} & \multicolumn{1}{c|}{\cellcolor{white}} & \multicolumn{1}{c|}{1.952}   & \multicolumn{1}{c|}{2.355}    & \multicolumn{1}{c|}{0.872}   & \multicolumn{1}{c|}{0.753}  \\ \cline{1-4} \cline{7-10}  \cline{13-16}
\rowcolor{lightgray}
\multicolumn{1}{|c|}{MS1M-RetinaFace} & \multicolumn{1}{c|}{iresnet\textbf{100}} & \multicolumn{1}{c|}{ArcFace} & \multicolumn{1}{c||}{FairFace} & \multicolumn{1}{c|}{\cellcolor{white}} & \multicolumn{1}{c|}{\cellcolor{white}} & \multicolumn{1}{c|}{0.889}   & \multicolumn{1}{c|}{0.954}    & \multicolumn{1}{c|}{0.951}   & \multicolumn{1}{c|}{0.927} & \multicolumn{1}{c|}{\cellcolor{white}} & \multicolumn{1}{c|}{\cellcolor{white}} & \multicolumn{1}{c|}{1.949}   & \multicolumn{1}{c|}{2.348}    & \multicolumn{1}{c|}{0.875}   & \multicolumn{1}{c|}{0.765}  \\ \cline{1-16} 
\end{tabular}}
\vspace{-7pt}
\end{table*}

\begin{table*}[h]
    \caption{Analysis of obfuscation-utility trade-off in facial recognition models using the iresnet-50 architecture. Performance is evaluated across varying information leakage weights $\alpha$, with significant differences between $\alpha = 0.1$ and $\alpha = 10$. Sensitive attributes considered are `Gender' and `Race' with a latent dimensionality of $d_{\mathbf{z}} = 256$. Notations: ``WF4M'' represents ``WebFace4M'', and ``MS1M-RF'' denotes ``MS1M-RetinaFace''.}
    \vspace{-9pt}
    \label{Table:PerformanceMetrics_FacialRecognitionModels_AfterPrivacyFunnel}
    \centering
    \resizebox{0.9\linewidth}{!}{%
    \begin{tabular}{c|cccccccccccc}
    \cline{2-13} 
    \multicolumn{1}{c||}{} & \multicolumn{6}{c|}{$\mathbf{S}$: Gender} & \multicolumn{6}{c|}{$\mathbf{S}$: Race} \\ \cline{2-13} 
\multicolumn{1}{c||}{}                                                                        & \multicolumn{3}{c|}{$\alpha = 0.1$}  & \multicolumn{3}{c|}{$\alpha = 10$}
& \multicolumn{3}{c|}{$\alpha = 0.1$}  & \multicolumn{3}{c|}{$\alpha = 10$}\\ \cline{1-13}
\rowcolor{headercolor}
\multicolumn{1}{|c||}{Face Recognition Model} 
& \multicolumn{1}{c|}{TMR@FMR=10e-1} & \multicolumn{1}{c|}{$\I(\mathbf{Z}; \mathbf{S})$} & \multicolumn{1}{c|}{Acc on $\mathbf{S}$} 
& \multicolumn{1}{c|}{TMR@FMR=10e-1} & \multicolumn{1}{c|}{$\I(\mathbf{Z}; \mathbf{S})$} & \multicolumn{1}{c|}{Acc on $\mathbf{S}$}
& \multicolumn{1}{c|}{TMR@FMR=10e-1} & \multicolumn{1}{c|}{$\I(\mathbf{Z}; \mathbf{S})$} & \multicolumn{1}{c|}{Acc on $\mathbf{S}$} 
& \multicolumn{1}{c|}{TMR@FMR=10e-1} & \multicolumn{1}{c|}{$\I(\mathbf{Z}; \mathbf{S})$} & \multicolumn{1}{c|}{Acc on $\mathbf{S}$}\\ \hline \cline{1-13}
\multicolumn{1}{|c||}{WF4M-i50-Ada-Morph} 
& \multicolumn{1}{c|}{93.60} & \multicolumn{1}{c|}{0.464} & \multicolumn{1}{c|}{0.992} 
& \multicolumn{1}{c|}{30.76} & \multicolumn{1}{c|}{0.388} & \multicolumn{1}{c|}{0.843}
& \multicolumn{1}{c|}{92.37} & \multicolumn{1}{c|}{0.628} & \multicolumn{1}{c|}{0.997} 
& \multicolumn{1}{c|}{30.03} & \multicolumn{1}{c|}{0.550} & \multicolumn{1}{c|}{0.857}
\\ \hline \cline{1-7} 
\multicolumn{1}{|c||}{MS1M-RF-i50-Arc-Morph} 
& \multicolumn{1}{c|}{94.05} & \multicolumn{1}{c|}{0.485} & \multicolumn{1}{c|}{0.992} 
& \multicolumn{1}{c|}{58.67} & \multicolumn{1}{c|}{0.335} & \multicolumn{1}{c|}{0.846}
& \multicolumn{1}{c|}{94.01} & \multicolumn{1}{c|}{0.635} & \multicolumn{1}{c|}{0.997} 
& \multicolumn{1}{c|}{58.34} & \multicolumn{1}{c|}{0.558} & \multicolumn{1}{c|}{0.868}
\\ \hline \cline{1-7} 
\multicolumn{1}{|c||}{WF4M-i50-Ada-FairFace} 
& \multicolumn{1}{c|}{94.83} & \multicolumn{1}{c|}{0.638} & \multicolumn{1}{c|}{0.925} 
& \multicolumn{1}{c|}{42.95} & \multicolumn{1}{c|}{0.367} & \multicolumn{1}{c|}{0.576}
& \multicolumn{1}{c|}{94.62} & \multicolumn{1}{c|}{0.866} & \multicolumn{1}{c|}{0.946} 
& \multicolumn{1}{c|}{42.13} & \multicolumn{1}{c|}{0.595} & \multicolumn{1}{c|}{0.756}
\\ \hline \cline{1-7} 
\multicolumn{1}{|c||}{MS1M-RF-i50-Arc-FairFace} 
& \multicolumn{1}{c|}{88.28} & \multicolumn{1}{c|}{0.636} & \multicolumn{1}{c|}{0.915} 
& \multicolumn{1}{c|}{59.91} & \multicolumn{1}{c|}{0.388} & \multicolumn{1}{c|}{0.598}
& \multicolumn{1}{c|}{95.57} & \multicolumn{1}{c|}{0.899} & \multicolumn{1}{c|}{0.947} 
& \multicolumn{1}{c|}{60.33} & \multicolumn{1}{c|}{0.608} & \multicolumn{1}{c|}{0.766}
\\ \hline \cline{1-7} 
\end{tabular}}
\vspace{-12pt}
\end{table*}

\vspace{-7pt}

\subsection{DVPF Objectives}

\vspace{-5pt}

\begin{figure}[!t]
\centering
\includegraphics[width=0.7\columnwidth]{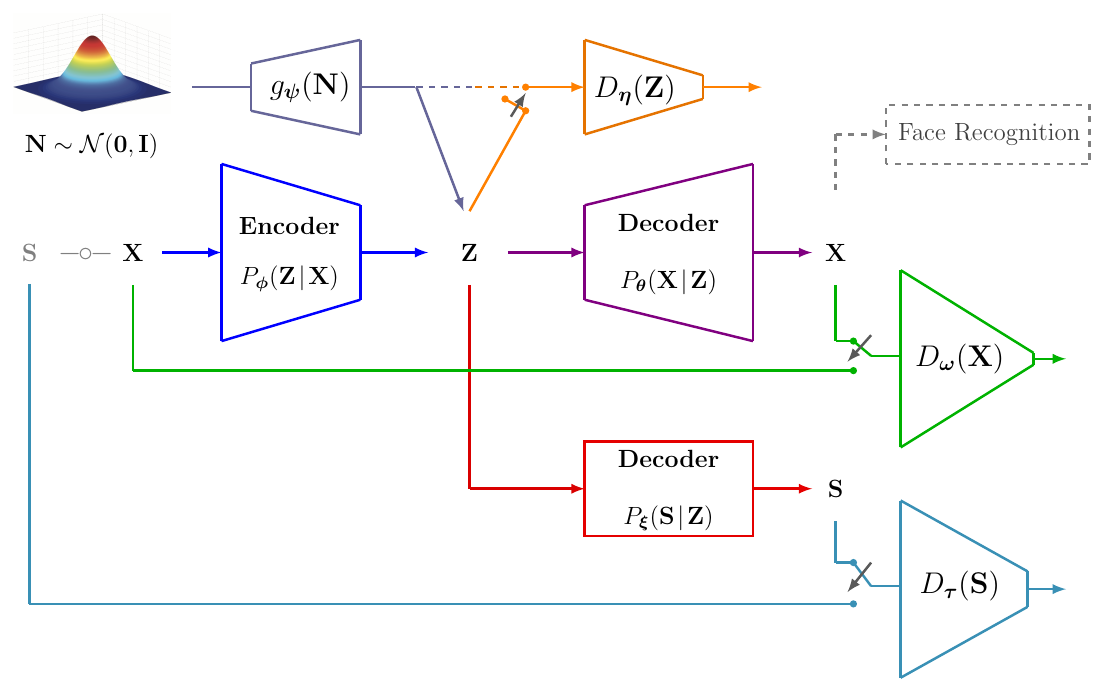}
\vspace{-10pt}
\caption{Training architecture associated with $\textsf{DVPF}\, (\textsf{P1})$.}
\label{Fig:DVPF_complete_LowerBound}
\vspace{-12pt} 
\end{figure}

Given \eqref{Eq:PF_LagrangianFunctional} and the parameterized approximations detailed earlier, the DVPF Lagrangian functional can be derived. 
Specifically, considering \eqref{Eq:I_SZ_phi_xi_SecondDecomposition} and \eqref{Eq:I_XZ_phi_theta}, we propose this objective:\vspace{-4pt}
%
%
%
\begin{align}\label{Eq:DVCLUB_Supervised_SLowerBound}
& (\textsf{P1})\! :  \; 
\mathcal{L}_{\mathrm{DVPF}} \left( \boldsymbol{\phi}, \boldsymbol{\theta} , \boldsymbol{\xi}, \alpha \right)  \coloneqq \\
& \overbrace{ 
- \H_{\boldsymbol{\phi}, \boldsymbol{\theta}} \left( \mathbf{X} \! \mid  \! \mathbf{Z} \right) - \D_{\mathrm{KL}} \left( P_{\mathsf{D}} (\mathbf{X}) \Vert P_{\boldsymbol{\theta}} (\mathbf{X}) \right)
}^{\textcolor{violet}{\mathrm{Information~Utility:}~  \I_{\boldsymbol{\phi}, \boldsymbol{\theta}}^{\mathrm{L}} \left( \mathbf{X}; \mathbf{Z} \right) }}
\nonumber \\
& -  \alpha \underbrace{ \Big( \! - \H_{\boldsymbol{\phi}, \boldsymbol{\xi}} \left( \mathbf{S} \! \mid  \! \mathbf{Z} \right) 
+ \H \left( P_{\mathbf{S}} \, \Vert \, P_{\boldsymbol{\xi}} (\mathbf{S}) \right) - \D_{\mathrm{KL}} \! \left( P_{\mathbf{S}} \Vert P_{\boldsymbol{\xi}} (\mathbf{S}) \right) \! \Big) }_{\textcolor{red}{\mathrm{Information~Leakage:}~\I_{\boldsymbol{\phi}, \boldsymbol{\xi}}  \left( \mathbf{S} ; \mathbf{Z} \right) }} . \nonumber 
\end{align}
Considering the upper bound \eqref{Eq:I_SZ_phi_Xi_UpperBound}, the corresponding objective~is:\vspace{-6pt}
\begin{multline}\label{Eq:DVCLUB_UnSupervised}
\!\!\!\!\!\! (\textsf{P2})\! :  \; 
\mathcal{L}_{\mathrm{DVPF}} \left( \boldsymbol{\phi}, \boldsymbol{\theta}, \boldsymbol{\psi} , \boldsymbol{\varphi}, \alpha \right)  \coloneqq  \\
 - \H_{\boldsymbol{\phi}, \boldsymbol{\theta}} \left( \mathbf{X} \! \mid  \! \mathbf{Z} \right) - \D_{\mathrm{KL}} \left( P_{\mathsf{D}} (\mathbf{X}) \, \Vert \, P_{\boldsymbol{\theta}} (\mathbf{X}) \right)
\\ \;\;\; -   \alpha 
 \underbrace{ \Big(  \I_{\boldsymbol{\phi}, \boldsymbol{\psi}} \left( \mathbf{X}; \mathbf{Z} \right) + \H_{\boldsymbol{\phi}, \boldsymbol{\varphi}}^{\mathrm{U}}   \left( \mathbf{X}  \mid \mathbf{S}, \mathbf{Z} \right) \Big) }_{\textcolor{red}{\mathrm{Information~Leakage}:~\I_{\boldsymbol{\phi}, \boldsymbol{\psi}, \boldsymbol{\varphi}}^{\mathrm{U}}   \left( \mathbf{S}; \mathbf{Z} \right)}}. \!\!\!
\end{multline}
\autoref{Fig:DVPF_complete_LowerBound} illustrates the training architecture for $(\textsf{P1})$. 
Due to space constraints, we present only the results for $(\textsf{P1})$.

\noindent
\textbf{Learning Procedure:}
The DVPF model $(\textsf{P1})$ is trained using alternating block coordinate descent across six steps:

\noindent
(1) {\small\textbf{\textsf{Train Encoder, Utility and Uncertainty Decoders}}}.\vspace{-5pt}
\begin{align}
     & \!\!\!\!\!\! \mathop{\max}_{\boldsymbol{\phi}, \boldsymbol{\theta}, \boldsymbol{\xi}} \; \mathbb{E}_{P_{\mathsf{D}}(\mathbf{X})} \left[ \mathbb{E}_{P_{\boldsymbol{\phi}} (\mathbf{Z} \mid \mathbf{X})} \left[ \log P_{\boldsymbol{\theta}} (\mathbf{X} \! \mid \! \mathbf{Z} ) \right]  \right]  \nonumber\\
    &\!\!\!\!\!\! -   \alpha \; \mathbb{E}_{P_{\mathbf{S}, \mathbf{X}}} \left[  \mathbb{E}_{P_{\boldsymbol{\phi}} \left( \mathbf{Z} \mid \mathbf{X} \right) } \left[ \log P_{\boldsymbol{\xi}}  \left( \mathbf{S} \! \mid \! \mathbf{Z} \right) \right] \right] 
    -  \alpha \; \mathbb{E}_{P_{\mathbf{S}}} \left[ \log P_{\boldsymbol{\xi}} (\mathbf{S}) \right]. \!\!\!\! \nonumber
\end{align}
\noindent
(2) {\small\textbf{\textsf{Train Latent Space Discriminator}}}.\vspace{-9pt}
    \begin{multline}
        \mathop{\min}_{\boldsymbol{\eta}}  \quad   \mathbb{E}_{P_{\mathsf{D}}(\mathbf{X})} \left[ \, \mathbb{E}_{P_{\boldsymbol{\phi}}(\mathbf{Z} \mid \mathbf{X})} \left[ - \log D_{\boldsymbol{\eta}} (\mathbf{Z}) \right] \, \right] \\+ \mathbb{E}_{Q_{\boldsymbol{\psi}} (\mathbf{Z})} \left[ \, - \log (1 - D_{\boldsymbol{\eta}} (\mathbf{Z})) \, \right]. \nonumber
    \end{multline}
\noindent
(3) {\small\textbf{\textsf{Train Encoder and Prior Distribution Generator Adversarially}}}.\vspace{-9pt}
    \begin{multline}
        \mathop{\max}_{\boldsymbol{\phi}, \boldsymbol{\psi}}  \quad   \mathbb{E}_{P_{\mathsf{D}}(\mathbf{X})} \left[ \, \mathbb{E}_{P_{\boldsymbol{\phi}}(\mathbf{Z} \mid \mathbf{X})} \left[ - \log D_{\boldsymbol{\eta}} (\mathbf{Z}) \right] \, \right] \\+ \mathbb{E}_{Q_{\boldsymbol{\psi}} (\mathbf{Z})} \left[ \, - \log (1 - D_{\boldsymbol{\eta}} (\mathbf{Z})) \, \right].\nonumber
    \end{multline}
\noindent
(4) {\small\textbf{\textsf{Train Utility Output Space Discriminator}}}.\vspace{-5pt}
\begin{eqnarray}\label{Eq:TrainVisibleSpaceDiscriminator}
    \mathop{\min}_{\boldsymbol{\omega}}   \mathbb{E}_{P_{\mathsf{D}}(\mathbf{X})} \! \left[ - \log D_{\boldsymbol{\omega}} (\mathbf{X})   \right] \! + \mathbb{E}_{Q_{\boldsymbol{\psi}} (\mathbf{Z})} \! \left[ - \log \left( 1 \! -\!  D_{\boldsymbol{\omega}} ( g_{\boldsymbol{\theta}}(\mathbf{Z} ) ) \right) \right].\nonumber
\end{eqnarray}
\noindent
(5) {\small\textbf{\textsf{Train Sensitive Attribute Class Discriminator}}}.\vspace{-4pt}
\begin{eqnarray}
    \mathop{\min}_{\boldsymbol{\tau}}  \,  \mathbb{E}_{P_{\mathbf{S}}} \left[ - \log D_{\boldsymbol{\tau}} (\mathbf{S}) \right] + \mathbb{E}_{Q_{\boldsymbol{\psi}} (\mathbf{Z})} \left[ - \log \left( 1 - D_{\boldsymbol{\tau}} ( g_{\boldsymbol{\xi}}(\mathbf{Z} ) ) \right) \right].\nonumber
\end{eqnarray}
\noindent
(6) {\small\textbf{\textsf{Train Prior Distribution Generator and Utility Decoder Adversarially}}}.\vspace{-9pt}
\begin{eqnarray}\label{TrainPriorDistributionGeneratorAndUtilityDecoderAdversarially}
    \mathop{\max}_{\boldsymbol{\psi}, \boldsymbol{\theta}}  \quad  \mathbb{E}_{Q_{\boldsymbol{\psi}} (\mathbf{Z})} \left[ \, - \log \left( 1 - D_{\boldsymbol{\omega}} ( \, g_{\boldsymbol{\theta}}(\mathbf{Z} ) \, ) \right) \, \right] .\nonumber
\end{eqnarray}

%

\vspace{-8pt}

\section{Experiments}
\label{Sec:Experiments}

\vspace{-8pt}

In this condensed study, we delve into the methodology of \textit{Embedding-Based Data Learning} for facial image analysis. We've excluded detailed results and discussions, such as the bounds of information leakage, various plots, and methodologies like \textit{Raw Data Transfer Learning with Fine-Tuning} and \textit{End-to-End Raw Data Scratch Learning}. Notably, the generative variational privacy funnel, vital for private synthetic data generation, will be extensively covered in our upcoming extended research version.

We consider the state-of-the-art Face Recognition (FR) backbones with three variants of IResNet \cite{resnet2016,arcface2019} architecture (IResNet100, IResNet50, and IResNet18). These architectures have been trained using either the MS1MV3 \cite{deng2019lightweight_ms1mv3} or WebFace4M/12M \cite{zhu2021webface260m} datasets.
For loss functions, ArcFace \cite{arcface2019} and AdaFace \cite{kim2022adaface} methods were employed. 
\autoref{Table:PerformanceMetrics_FacialRecognitionModels} depicts the Shannon entropy, estimated mutual information between the extracted embeddings $\mathbf{X} \in \mathbb{R}^{512}$ and sensitive attributes $\mathbf{S}$, and accuracy of recognition of $\mathbf{S}$, for test and train sets, before applying our DVPF model. 
For the training phase, we utilized pre-trained models sourced from the aforementioned studies. All input images underwent a standardized pre-processing routine, encompassing alignment, scaling, and normalization. This was in accordance with the specifications of the pre-trained models. We then trained our networks using the Morph dataset \cite{morph1} and FairFace \cite{karkkainenfairface}, focusing on different demographic group combinations such as race and gender.
A close proximity between $\I( \mathbf{X}; \mathbf{S})$ and entropy $\H( \mathbf{S} )$ indicates that the embeddings considerably mitigate label uncertainty. Given $\I( \mathbf{X}; \mathbf{S}) = \H( \mathbf{X} ) + \H( \mathbf{S} ) - \H( \mathbf{X}, \mathbf{S} ) $, mutual information serves as a measure of the reduced joint uncertainty about $\mathbf{X}$ and $\mathbf{S}$. It's pivotal to note that $\I( \mathbf{X}; \mathbf{S}) \leq \min \left( \H( \mathbf{X} ) , \H( \mathbf{S} )\right)$. 
For the Morph/FairFace datasets, the entropy of sensitive attributes (gender or race) remains consistent across both train/test sets and differing FR model embeddings, emphasizing the same dataset usage throughout experiments. Both Morph and FairFace datasets, featuring `male' and `female' gender labels, attain a maximum entropy of $ \log_2(2) = 1 $. The Morph dataset, with four distinct race labels, reaches a maximum entropy of $ \log_2(4) = 2 $, while the FairFace dataset, with six race labels, tops at $ \log_2(6) = 2.585 $. 
Within Morph, the mutual information for gender mirrors its entropy, suggesting notable preservation of sensitive information in the embeddings. However, for race, values of approximately $0.92\text{-}0.93$ underscore an imbalanced label distribution, as they don't reach the theoretical $ \log_2(4) = 2 $. 
In contrast, the FairFace dataset displays near-maximal entropies for race ($\sim 2.517$ relative to a potential $2.585$) and gender ($\sim 0.999$ compared to an ideal $1$), illustrating well-balanced racial and gender label distributions.

\vspace{-4pt}
 
We applied our DVPF model to the embeddings obtained from the FR models referenced in \autoref{Table:PerformanceMetrics_FacialRecognitionModels}. For the accuracy evaluation of our DVPF model within the facial recognition domain, we utilized the challenging IJB-C test dataset \cite{ijbc} as our benchmark. The assessment was initiated with the pre-trained backbones, followed by our DVPF model, which was developed using embeddings from these pre-trained structures. Given space constraints and the consistent performance observed across various IResNet architectures, we present results specific to the IResNet50.

In \autoref{Table:PerformanceMetrics_FacialRecognitionModels_AfterPrivacyFunnel}, we precisely quantify the disclosed information leakage, represented as $\I(\mathbf{S}; \mathbf{Z})$. Additionally, we provide a detailed account of the accuracy achieved in recognizing sensitive attributes from the disclosed representation $\mathbf{Z} \in \mathbb{R}^{256}$, utilizing the support vector classifier optimization. These evaluations are based on test sets derived from either the Morph or FairFace datasets.
Moreover, we detail the True Match Rate ($\mathsf{TMR}$) for our models. It's imperative to note that all these evaluations are systematically benchmarked against a predetermined False Match Rate ($\mathsf{FMR}$) of $10^{-1}$.
When subjecting the `WF4M-i50-Ada' model to evaluation against the IJB-C dataset---prior to the DVPF model's integration---a $\mathsf{TMR}$ of $\mathsf{99.40\%}$ at $\mathsf{FMR=10e-1}$ was observed. Similarly, for the `MS1M-RF-i50-Arc' configuration, a $\mathsf{TMR}$ of $\mathsf{99.58\%}$ was observed on the IJB-C dataset before the integration of the DVPF model, with measurements anchored to the same $\mathsf{FMR}$. Consistent with our expectations, as $\alpha$ increases towards infinity ($\alpha \rightarrow \infty$), the information leakage $\I (\mathbf{S}; \mathbf{Z})$ decreases to zero. At the same time, the recognition accuracy for the sensitive attribute $\mathbf{S}$ approaches $0.5$, indicative of random guessing.

\vspace{-8pt}

%
%
%
\section{Conclusion}
\label{Sec:Conclusions}
\vspace{-8pt}

In this study, we integrate the privacy funnel model for privacy-preserving deep learning, bridging information-theoretic privacy and representation learning. 
Applied to the state-of-the-art face recognition models, our approach underscores the balance between information obfuscation and utility. The model enhances data protection in discriminative and generative contexts, with an accompanying reproducible software package facilitating further research exploration and adoption.


\clearpage

\bibliographystyle{IEEEbib}
\bibliography{references}

\end{document}

%% file: preamble.tex
\usepackage[utf8]{inputenc} 
\usepackage[T1]{fontenc}    
\usepackage{hyperref}       
\hypersetup{
    colorlinks,
    linkcolor={red!50!black},
    citecolor={blue!50!black},
    urlcolor={black}
}
\usepackage{url}            
\usepackage{booktabs}       
\usepackage{amsfonts}       
\usepackage{nicefrac}       
\usepackage{microtype}      
\usepackage{doi}

\usepackage{algorithm}
\usepackage{algpseudocode}
\usepackage{graphicx}
\usepackage{textcomp}
\usepackage{xcolor}
\def\BibTeX{{\rm B\kern-.05em{\sc i\kern-.025em b}\kern-.08em
    T\kern-.1667em\lower.7ex\hbox{E}\kern-.125emX}}

\usepackage{amssymb}
\usepackage{amssymb}
\usepackage{amsbsy}
\usepackage{amsthm, mathrsfs, amsmath}
\usepackage{longtable}
\usepackage{stackengine}
\usepackage{mathtools}
\usepackage{stmaryrd}
\usepackage{relsize}
\usepackage{multirow}
\usepackage{bm}
\usepackage{subcaption}
\usepackage{caption}
\usepackage{dsfont}
\usepackage{cite}
\usepackage{color}   
\usepackage{epsfig}
\usepackage{xfrac}
\usepackage{setspace}

\usepackage{array}
\usepackage{tabularx}

\usepackage{cuted}
\usepackage{multicol}
\usepackage{lipsum}

\usepackage{float}

\usepackage{hyperref}

\usepackage{amstext}
\usepackage{amssymb,amsmath,amsthm,enumitem}
\usepackage{sansmath}

\usepackage{mathrsfs}
\DeclareMathAlphabet{\mathpzc}{OT1}{pzc}{m}{it}

\def \H {\mathrm{H}} 
\def \I {\mathrm{I}} 
\def \D {\mathrm{D}} 

\theoremstyle{definition}

\def\markov{\hbox{$\--$}\kern-1.5pt\hbox{$\circ$}\kern-1.5pt\hbox{$\--$}}

\makeatletter
\newcommand*{\centernot}{%
  \mathpalette\@centernot
}
\def\@centernot#1#2{%
  \mathrel{%
    \rlap{%
      \settowidth\dimen@{$\m@th#1{#2}$}%
      \kern.5\dimen@
      \settowidth\dimen@{$\m@th#1=$}%
      \kern-.5\dimen@
      $\m@th#1\not$%
    }%
    {#2}%
  }%
}
\makeatother

\usepackage{makecell}

\usepackage{footmisc}

\usepackage{colortbl}

\definecolor{headercolor}{gray}{0.82} 
\definecolor{zebracolor}{gray}{0.95}  
\definecolor{lightgray}{gray}{0.93}

\usepackage{tikz}
\usetikzlibrary{calc}